\documentclass{article}
\pdfoutput=1

\PassOptionsToPackage{numbers, square,sort&compress}{natbib}
\usepackage[final]{nips_2016}

\usepackage[utf8]{inputenc} 
\usepackage[T1]{fontenc}    
\usepackage{hyperref}       
\usepackage{url}            
\usepackage{booktabs}       
\usepackage{amsfonts}       
\usepackage{nicefrac}       
\usepackage{microtype}      
\usepackage{cite}
\usepackage{amsmath}
\usepackage[disable]{todonotes}
\usepackage{subcaption}
\usepackage{tabularx}

\usepackage[nolist,nohyperlinks]{acronym}
\acrodef{DNN}{deep neural network}
\acrodef{SQNR}{signal-to-quantization noise ratio}
\acrodef{SC}{stochastic computing}
\acrodef{LFSR}{linear feedback shift register}
\acrodef{ReLU}{rectified linear unit}
\acrodef{BNN}{bitwise neural network}
\acrodef{mux}{multiplexer}
\acrodef{MAC}{multiply accumulate}
\acrodef{CU}{computational units}
\acrodef{MSB}{most significant bit}
\acrodef{GTSRB}{German Traffic Sign Recognition Benchmark}
\acrodef{BN}{batch normalization}
\acrodef{PRBS}{pseudo random bitstream}
\acrodef{GPU}{graphics processing unit}

\title{Efficient Stochastic Inference of Bitwise Deep Neural Networks}

\author{
  Sebastian~Vogel\thanks{These authors contributed equally to this work.} \\
  Robert Bosch GmbH \\
  Corporate Research Campus\\
  71272 Renningen, Germany \\
  \texttt{sebastian.vogel@de.bosch.com}
  \And
  Christoph~Schorn\footnotemark[1] \\
  Robert Bosch GmbH \\
  Corporate Research Campus \\
  71272 Renningen, Germany \\
  \texttt{christoph.schorn@de.bosch.com}
  \And
  Andre~Guntoro \\
  Robert Bosch GmbH \\
  Corporate Research Campus \\
  71272 Renningen, Germany \\
  \texttt{andre.guntoro@de.bosch.com}
  \And
  Gerd~Ascheid\thanks{Professor Gerd Ascheid is Senior Member IEEE.} \\
  Institute for Communication\\ Technologies and Embedded Systems \\
  RWTH Aachen University, Germany\\
  \texttt{ascheid@ice.rwth-aachen.de} \\
}

\begin{document}

\maketitle

\begin{abstract}
Recently published methods enable training of bitwise neural networks which allow reduced representation of down to a single bit per weight. We present a method that exploits ensemble decisions based on multiple stochastically sampled network models to increase performance figures of bitwise neural networks in terms of classification accuracy at inference. Our experiments with the CIFAR\hbox{-}10 and GTSRB datasets show that the performance of such network ensembles surpasses the performance of the high-precision base model. With this technique we achieve 5.81\% best classification error on CIFAR\hbox{-}10 test set using bitwise networks. Concerning inference on embedded systems we evaluate these bitwise networks using a hardware efficient stochastic rounding procedure. Our work contributes to efficient embedded bitwise neural networks. 
\end{abstract}

\section{Introduction}

Research results in recent years have shown tremendous advances in solving complex problems using deep learning approaches. Especially classification tasks based on image data have been a major target for \acp{DNN} \citep{Kriz12, He15}. A challenge for leveraging the strengths of deep learning methods in embedded systems is their massive computational cost. Even relatively small \acp{DNN} often require millions of parameters and billions of operations for performing a single classification.  
Model compression approaches can help to relax memory requirements as well as to reduce the number of required operations of \acp{DNN}. While some approaches consider special network topologies \citep{He15, Iand16}, another stream of research focuses on precision reduction of the model parameters. Recent publications of \acp{BNN} have shown that network weights and activations can be reduced from a high-precision floating-point down to a binary representation, while maintaining classification accuracy on benchmark datasets \citep{Cour16}. 
Stochastic projection of the network weights during training is a key component that enables this strong quantization. Studies which employed this training method have so far only analyzed deterministic projections during test-time \citep{Cour15, Cour16, Mero16}. 

With techniques presented in this paper, we contribute to stochastic inference of bitwise neural networks on hardware. We show that stochastic rounding at test-time improves classification accuracy of networks that were trained with stochastic weight projections (Section~\ref{stochastic_inference}). Furthermore, we present a method which efficiently realizes stochastic rounding of network weights in a dedicated hardware accelerator (Section~\ref{efficient_stochastic_rounding_in_hardware}). We start off with a brief review of the literature on weight discretization (Section~\ref{related_work}).
\section{Related Work}\label{related_work}

Some recent studies have shown that weights (and activations) of \acp{DNN} can be discretized to a very low number of quantization levels while maintaining high classification performance \citep{Anwa15, Cour15, Cour16, Hwan14, Kim14, Mero16, Rast16}. They employ a method which has already been sketched out by \citep{Fies90}. For each iteration of the back-propagation learning algorithm the high-precision weights of the network are projected to discretized values. The discrete weights are used to compute gradient descent based weight updates, which are then \emph{applied to the high-precision weights}. This method can be used either as a fine-tuning step for several epochs after regular training \citep{Anwa15, Hwan14, Kim14} or from the beginning of the training~\citep{Cour15, Cour16, Mero16, Rast16}. \citep{Cour15} has recently introduced clipping followed by stochastic rounding as a method for projecting high-precision to binary (-1, +1) weights. Before, \citep{Gupt15} used a similar method but with a relatively large number of discretization levels and presented a neural network hardware accelerator using multiply-accumulate-units for stochastic rounding. Instead, we present a method avoiding multipliers.

\section{Stochastic Inference}\label{stochastic_inference}
Our methods are based on neural networks which are trained with stochastic weight projections. In this section, we show that by applying these projections at test-time, a stochastic ensemble of \acp{BNN} can be created whose aggregated classification performance surpasses that of the underlying high-precision floating-point model, while maintaining the benefits of bitwise and multiplierless computations. 

\subsection{Stochastic Network Ensembles}\label{stochastic_network_ensembles}
We employ the method introduced in~\citep{Cour15} during training \emph{and} inference. Depending on the number of discrete values we speak of \textit{binary} or \textit{ternary} network weights. Clipping limits the numerical range of the weights to the interval $\lbrack-1,1\rbrack$ and the projection $W\mapsto W^d$ is done by stochastic rounding:
\begin{center}\begin{equation}\label{eq:sround}sround(w)=\begin{cases}\lceil w\rceil, & \mbox{with probability}~p=\left| \frac{\lfloor w\rfloor-w}{\lfloor w\rfloor-\lceil w\rceil}\right| \\ \lfloor w\rfloor, & \mbox{with probability}~1-p=\left| \frac{\lceil w\rceil-w}{\lfloor w\rfloor-\lceil w\rceil}\right|\end{cases}.\end{equation}\end{center}
\begin{figure}[!htb]
\centering
\includegraphics[width=0.75\linewidth]{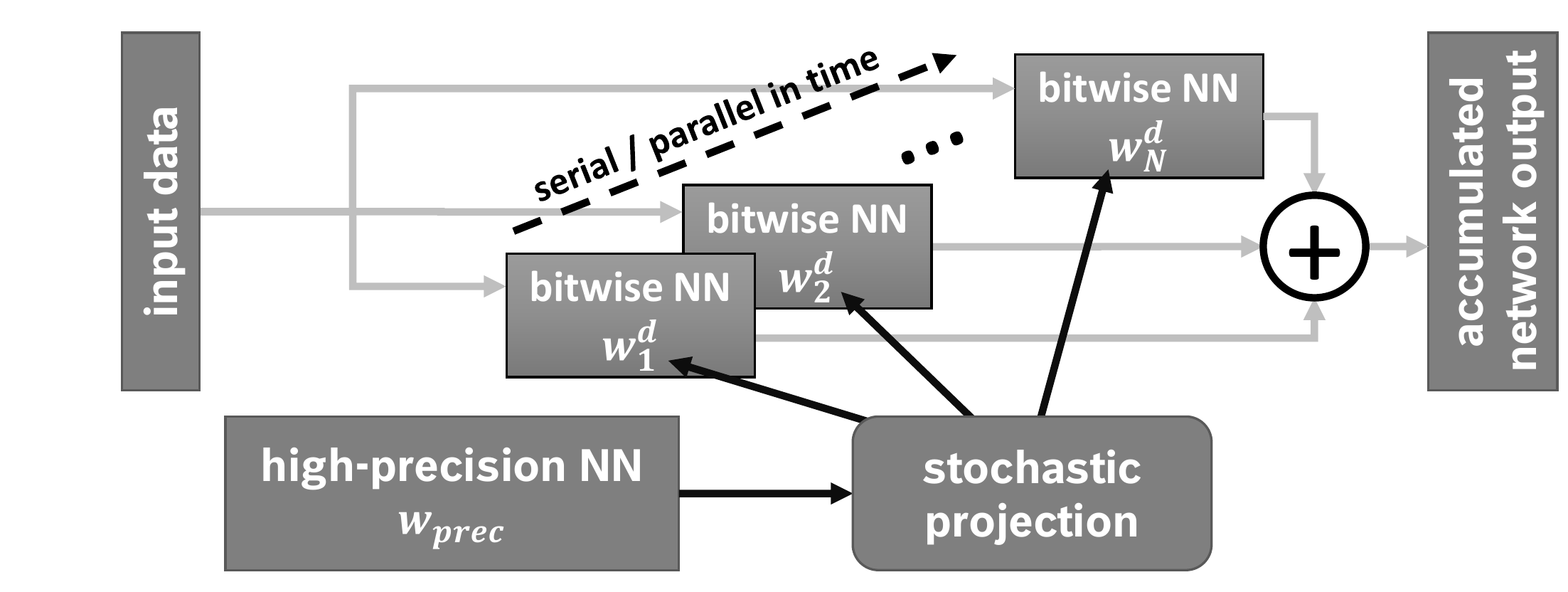}
\caption{Based on a high-precision network, an ensemble of networks is created. The outputs of the ensemble members are used in conjunction. }
\label{fig:ensemble_networks}
\end{figure}
Best test-time results in \citep{Cour15} were achieved with the high-precision neural network parameters $W$. However, discretized values are much better suited for dedicated hardware accelerators, which is why we investigate inference based on $W^d$. One approach is to perform inference at test-time with the same weight discretization projections as in the training procedure. The reasoning behind this is that the network has been optimized for these projections when minimizing the loss function. With Eqn.~\eqref{eq:sround} as projection function, experiments show a high variance in classification accuracy when the projection is performed only once. Ensembles of classifiers can be used to lower the classification variance of the aggregated classification decision. Using multiple stochastic projections $W\mapsto W^d$ we sample different versions of our neural network and combine their outputs as visualized in Figure~\ref{fig:ensemble_networks}. The ensemble classification decision is then taken based on this accumulated network output.

\subsection{Experimental Results}\label{network_ensembles}
\begin{figure}[!htb]
\centering
\caption{We evaluated ensembles of networks which were generated by stochastically projecting the high-precision model parameters to ternary values.}
\label{fig:CIFAR}
\begin{subfigure}{.49\textwidth}
  \centering
  \includegraphics[width=1.\linewidth]{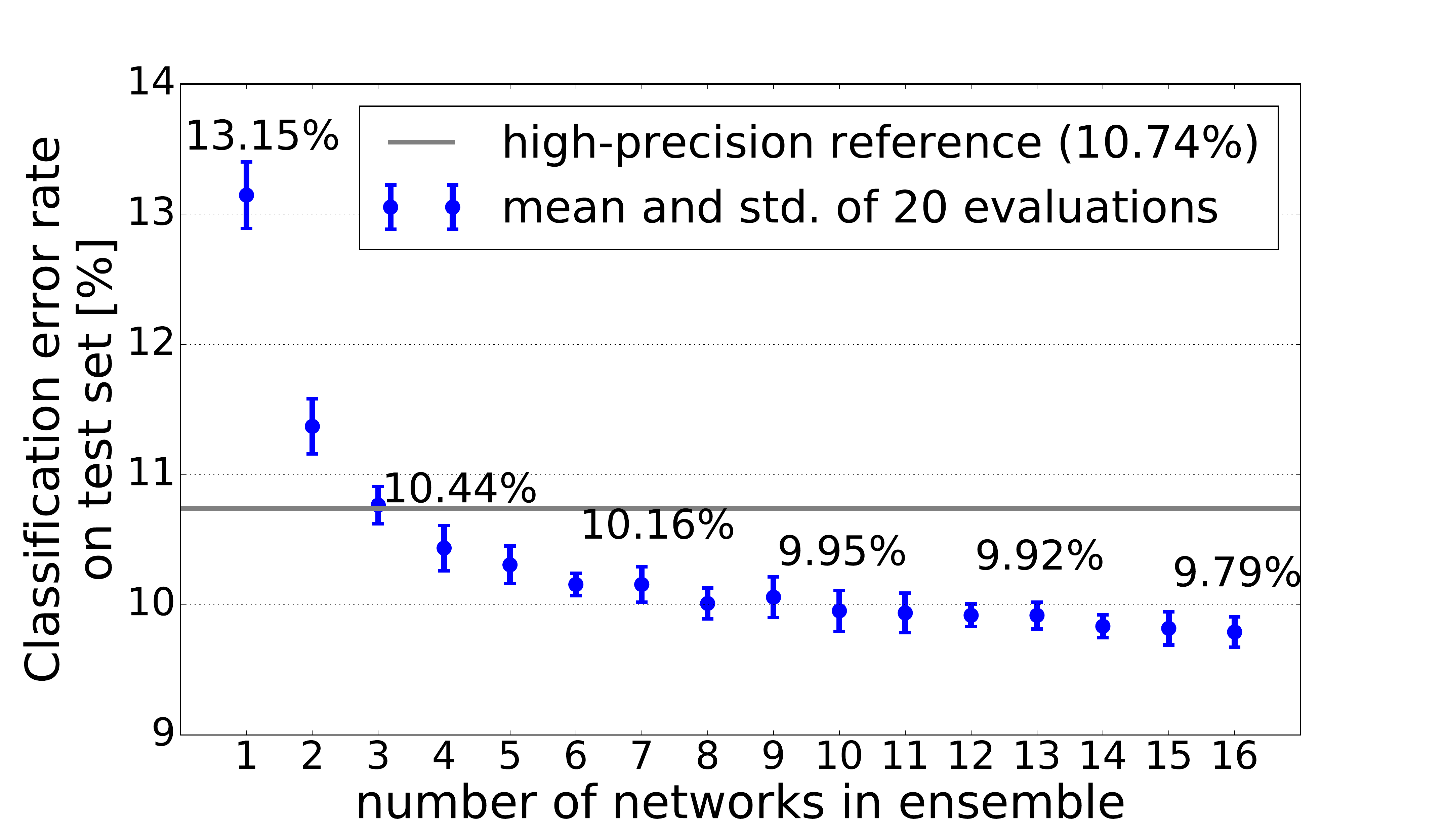}
  \caption{The network has $sign$ activation. Our best result was 9.41\% with an ensemble of 23 networks.}
  \label{fig:CIFARhtanh}
\end{subfigure}
\hspace{1mm}
\begin{subfigure}{.49\textwidth}
  \centering
  \includegraphics[width=1.\linewidth]{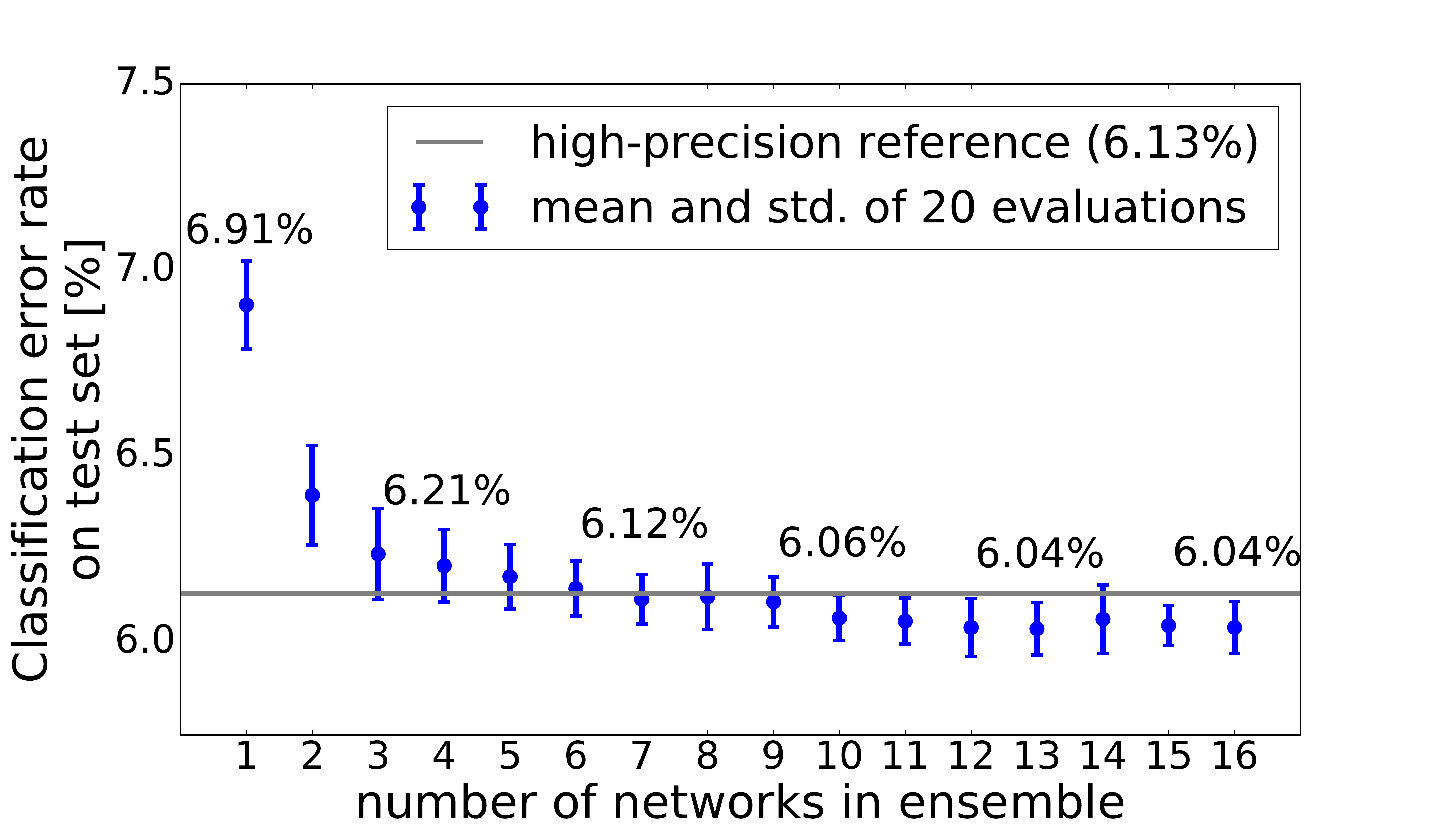}
  \caption{The network uses \ac{ReLU} activation. The best result of 5.81\% was achieved for an ensemble of 29 networks.}
  \label{fig:CIFARReLU}
\end{subfigure}
\end{figure}

For the first evaluation of our method, we train a ConvNet on the CIFAR\hbox{-}10 classification dataset~\citep{Kriz09}, which contains 60\,000 images in 32$\times$32 pixel RGB resolution and 10 different classes. We use the setup described in~\citep{Cour15} for training, but with $sign$\footnote{$sign(x)$: $1~\text{for}\geq0$, $-1$ otherwise.} activation function as in~\citep{Cour16} and stochastic ternary weights. The network structure is 128C3--128C3--MP2--256C3--256C3--MP2--512C3--512C3--MP2--1024FC--1024FC--10SVM\footnote{Preceding numbers indicate the number of channels, C3 denotes a convolution layer with 3$\times$3 kernel, MP2 abbreviates spatial max-pooling with a receptive field of 2$\times$2, FC stands for fully connected layers and SVM for a square hinge loss output layer.}. 
After training the model for 500 epochs with hyperparameters from~\citep{Cour15} and without any preprocessing or augmentations on the dataset, we select high-precision model parameters which have the lowest error on the validation set. These weights are used to generate multiple instances of the network by rounding the weights stochastically to ternary values (see Section~\ref{stochastic_network_ensembles}). Classification error rates on the CIFAR\hbox{-}10 test set based on the ensemble decision for different accumulation lengths, i.\,e. numbers of ensemble members, are plotted in Figure~\ref{fig:CIFARhtanh}. Since classification results are not deterministic in this case, we run the whole experiment 20$\times$ and provide mean and standard deviation. In our experiment, a stochastic \ac{BNN} ensemble with at least four members always performs better than the floating-point reference model, which achieves a classification error of 10.74\%.

Better classification results can be achieved when the same network is trained with \ac{ReLU} activation function, binary projections, global contrast normalization and ZCA whitening, as well as augmentations on the training data. We apply a commonly used simple data augmentation method \citep{Huan16}, consisting of a random translation of up to 4 pixels in the image plane and a random flip around the vertical axis. Classification results for this setup using ternary projections at test-time are shown in Figure \ref{fig:CIFARReLU}. The best result of 5.81\% was reached with an ensemble of 29 networks. To the best of our knowledge we are the first to report a classification error of less than 6\% on the CIFAR\hbox{-}10 benchmark using \aclp{BNN}.

In addition, we test our method on the \acl{GTSRB} dataset~\citep{Stal11}. The resulting high-precision network with $sign$ activation leads to 2.19\% classification error. For 20 evaluations, a single projected bitwise network results in 2.73\% mean error rate (0.092\%~std.) whereas ensembles of 11 networks reach 1.79\% mean error rate (0.042\%~std.). The best result of 1.63\% was achieved with 16 ensemble members.

Interestingly, the mean performance of discretized ensembles reach better classification results than the high-precision base model. We believe that due to the gradient descent optimization of the loss function which is evaluated for discrete values, best results are achieved with projected versions of the base model.
\section{Efficient Stochastic Rounding in Hardware}\label{efficient_stochastic_rounding_in_hardware}
In order to fully exploit the performance of \aclp{BNN} in terms of accuracy, the \ac{BNN} needs to be evaluated more than once and therefore an efficient integration of a stochastic rounding engine is necessary. Based on the publications~\citep{Bade94b} and~\citep{Bade94}, a simple multiplexer can be used to perform $sround(x)$ (see Eqn.~\eqref{eq:sround}). Assuming the probability of the select signal $sel$ of an N-to-1 multiplexer to route signal $in_i\in\{0,1\}$ to the output is equally distributed, the probability of the output signal $out$ being 1 can be written as 
\begin{center}\begin{equation}P(out=1)=\sum_{i=1}^N in_i P(sel=i)=\sum_{i=1}^N in_i \frac{1}{N}.\end{equation}\end{center}
Hence, the probability $P(out=1)$ is determined by the number of ones at the input $in$. However, if the probability function $P(sel=i)$ is chosen to be
\begin{center}\begin{equation}\label{eq:select_sig}P(sel=i)=\frac{2^{i-1}}{2^N-1},\end{equation}\end{center}
the probability $P(out=1)$ is directly related to the input $in$. Additionally, considering $in$ as a binary coded\footnote{$in_N$ corresponds to the \ac{MSB}.} fractional number $\in\left[0,1\right)$ then $P(out=1)\approx in$ with a maximum error of $\frac{1}{2^N}$. In order to use this technique in hardware, the corresponding signal for $sel$ has to be generated by individual select wires $sel_j$. Whereas~\citep{Bade94b} considers the $N$ equations~\eqref{eq:select_sig} as an overdetermined problem and proposes a numerical solution, we present an analytic solution to the problem. There are $log_2(N)$ individual select bits $sel_j$ with
\begin{equation}\label{eq:select_prob}\begin{aligned}P(sel_j=1)=\frac{2^{2^{j-1}}}{2^{2^{j-1}}+1}&,~~P(sel_j=0)=\frac{1}{2^{2^{j-1}}+1} \\ 
\Rightarrow\prod_{j=1}^{log_2(N)} P(sel_j)=P(sel),~~\text{be}&\text{cause}~~\prod_{k=1}^{log_2(M)}\left(2^{2^{k-1}}+1\right)=2^M-1.\end{aligned}\end{equation}
Bitstreams for $sel_j$ with the corresponding frequencies can be generated using a \ac{LFSR} in combination with Daalen modulators~\citep{Daal93}.

In order to verify the concept of stochastic rounding engines for neural networks using the method presented above, we evaluated the network for road sign recognition with weights stochastically projected in hardware. The results presented in Section~\ref{network_ensembles} have been reproduced using this approach. To take a potential hardware parallelization into consideration, we also performed projections in parallel over the dimension of output features. 
As the generation of random bitstreams using \ac{LFSR}s is expensive in terms of energy and hardware resources, we evaluated the classification performance when using a single \ac{PRBS} generator to provide the same select signal for all stochastic rounders (i.e. multiplexers) in the network. We found that relying on a single \ac{PRBS} generator retains mean classification accuracy. Moreover, the mean network performance is preserved when only a single \ac{LFSR} is used to generate a random base bitstream which is then subject to different modulations~\citep{Daal93} to generate \ac{PRBS} with appropriate frequencies of 1's (see Eqn.~\eqref{eq:select_prob}).
\section{Conclusion and Outlook}\label{conclusion}
We investigated \aclp{BNN} with stochastically projected weights during inference. Results show that an ensemble-based decision of multiple versions of such a \ac{BNN} enhances performance compared to the inference based on the high-precision shadow weights. Furthermore, we presented a hardware efficient stochastic rounding procedure for the first time used on bitwise \ac{DNN}s. Our results show that this technique can be used for test-time inference enabling efficient hardware implementation in embedded systems. 

The methods proposed in \citep{Cour15} and \citep{Cour16} rely on stochastic projections during training. Future research will investigate the integration of our generalized form of stochastic rounding into the training process.



\bibliographystyle{csplain}
\bibliography{literature}

\end{document}